\documentclass[runningheads,orivec]{llncs}

\usepackage[T1]{fontenc}

\usepackage[backend=biber,style=lncs,sorting=none]{biblatex}
\usepackage{hyperref}
\usepackage{amsmath,amssymb,amsfonts}
\usepackage{algorithmic}
\usepackage{graphicx}
\usepackage{textcomp}
\usepackage{array} 
\usepackage{multirow}
\graphicspath{{./figures/}}
\def\BibTeX{{\rm B\kern-.05em{\sc i\kern-.025em b}\kern-.08em
    T\kern-.1667em\lower.7ex\hbox{E}\kern-.125emX}}

\begin{document}
\title{Reinforcement Learning-Based Optimization of CT Acquisition and Reconstruction Parameters Through Virtual Imaging Trials}

\titlerunning{Reinforcement Learning-Based Optimization of CT Parameters}

\author{David~Fenwick\inst{1} \and Navid~NaderiAlizadeh\inst{1} \and Vahid~Tarokh\inst{1} \and Nicholas~Felice\inst{1} \and Darin~Clark\inst{1} \and Jayasai~Rajagopal\inst{2} \and Anuj~Kapadia\inst{2} \and Benjamin~Wildman-Tobriner\inst{1} \and Ehsan~Samei\inst{1} \and Ehsan~Abadi\inst{1}}

\authorrunning{D. Fenwick et al.}

\institute{Duke University, Durham, NC 27705 USA \and Oak Ridge National Laboratory, Oak Ridge, TN 37830 USA}

\maketitle
\begin{abstract}
Protocol optimization is critical in Computed Tomography (CT) to achieve high diagnostic image quality while minimizing radiation dose. However, due to the complex interdependencies among CT acquisition and reconstruction parameters, traditional optimization methods rely on exhaustive testing of combinations of these parameters, which is often impractical. This study introduces a novel methodology that combines virtual imaging tools with reinforcement learning to optimize CT protocols more efficiently. Human models with liver lesions were imaged using a validated CT simulator and reconstructed with a novel CT reconstruction toolkit. The optimization parameter space included tube voltage, tube current, reconstruction kernel, slice thickness, and pixel size. The optimization process was performed using a Proximal Policy Optimization (PPO) agent, which was trained to maximize an image quality objective, specifically the detectability index (\textbf{\(d'\)}) of liver lesions in the reconstructed images. Optimization performance was compared against an exhaustive search performed on a supercomputer. The proposed reinforcement learning approach achieved the global maximum \textbf{\(d'\)} across test cases while requiring 79.7\% fewer steps than the exhaustive search, demonstrating both accuracy and computational efficiency. The proposed framework is flexible and can accommodate various image quality objectives. The findings highlight the potential of integrating virtual imaging tools with reinforcement learning for CT protocol management.
\end{abstract}


\section{Introduction}
\label{sec:introduction}
Computed Tomography (CT) protocols include a set of acquisition and reconstruction parameters that collectively define how CT images are obtained and processed \cite{RN1}. These protocols must be optimized to achieve high diagnostic image quality metrics while keeping radiation dose as low as possible. A conventional approach to CT optimization involves acquiring images across a wide range of parameter settings and evaluating their effects on diagnostic image quality. However, such repetitive experiments are impractical in human subjects due to radiation exposure risks. Alternatively, these repeated scans can be acquired and analyzed using physical phantoms \cite{RN2,RN3}. While this approach is safer and more practical, physical phantoms lack the anatomical and pathological attributes of real patient populations, limiting their clinical applicability. Furthermore, the number of possible parameter combinations often makes it infeasible to test all configurations exhaustively. This leads to limited sampling of the parameter space and increases the risk of converging to sub-optimal protocols. 

These limitations can be addressed using realistic virtual imaging trials (VITs) \cite{RN14}, where a traditional imaging trial is simulated using validated human and scanner models. VITs enable controlled imaging experiments across a wide range of imaging conditions without exposing patients to radiation, while also providing anatomical and physiological ground-truth data. Due to these advantages, VITs have been widely used in recent years for image quality assessment and optimization \cite{RN11,RN12,RN13}. 

For instance, Barufaldi \textit{et. al.,} used VITs to evaluate and optimize a prototype for simultaneous Digital Breast Tomosynthesis (DBT) and Mechanical Imaging \cite{RN11}. Similarly, Vancoille \textit{et. al.,} used VITs to study microcalcification detectability in DBT and synthetic 2D mammography across different acquisition parameters \cite{RN12}. In another study, Abadi \textit{et. al.,} employed a VIT approach to test various beam collimation and pitch settings to determine their impact on image quality attributes for CT images with respiratory and cardiac motion \cite{RN13}. These studies demonstrate the utility of performing optimization tasks using a VIT approach. 

While these efforts have provided valuable insights, they have been limited by the need to study a reduced set of imaging parameters rather than conducting an exhaustive search, primarily due to the high computational demands associated with acquiring and reconstructing images across a large, complex parameter space. The purpose of this study is to address this limitation by developing an optimization framework that combines VITs with reinforcement learning \cite{RN35,RN36}, enabling efficient and accurate CT protocol optimization beyond limited parameter sampling. The proposed framework is benchmarked for its effectiveness in optimizing CT protocols for liver lesion detection. 

This paper is an expanded version of our previous conference publication \cite{fenwick2025}. Compared to the conference proceeding, this paper offers a more detailed and comprehensive description of the framework for task-based CT protocol optimization.  Additional experiments have been conducted, incorporating a patient cohort with greater variability in body habitus and lesion characteristics. Furthermore, we have expanded our quantitative analysis of image quality metrics and examined their relationships to acquisition and reconstruction parameters in greater depth.

\section{Methods}
\label{sec:methods}
\subsection{Study Framework}

\begin{figure*}[t]
    \centering
    \includegraphics[width = \textwidth, height = 0.3\textheight]{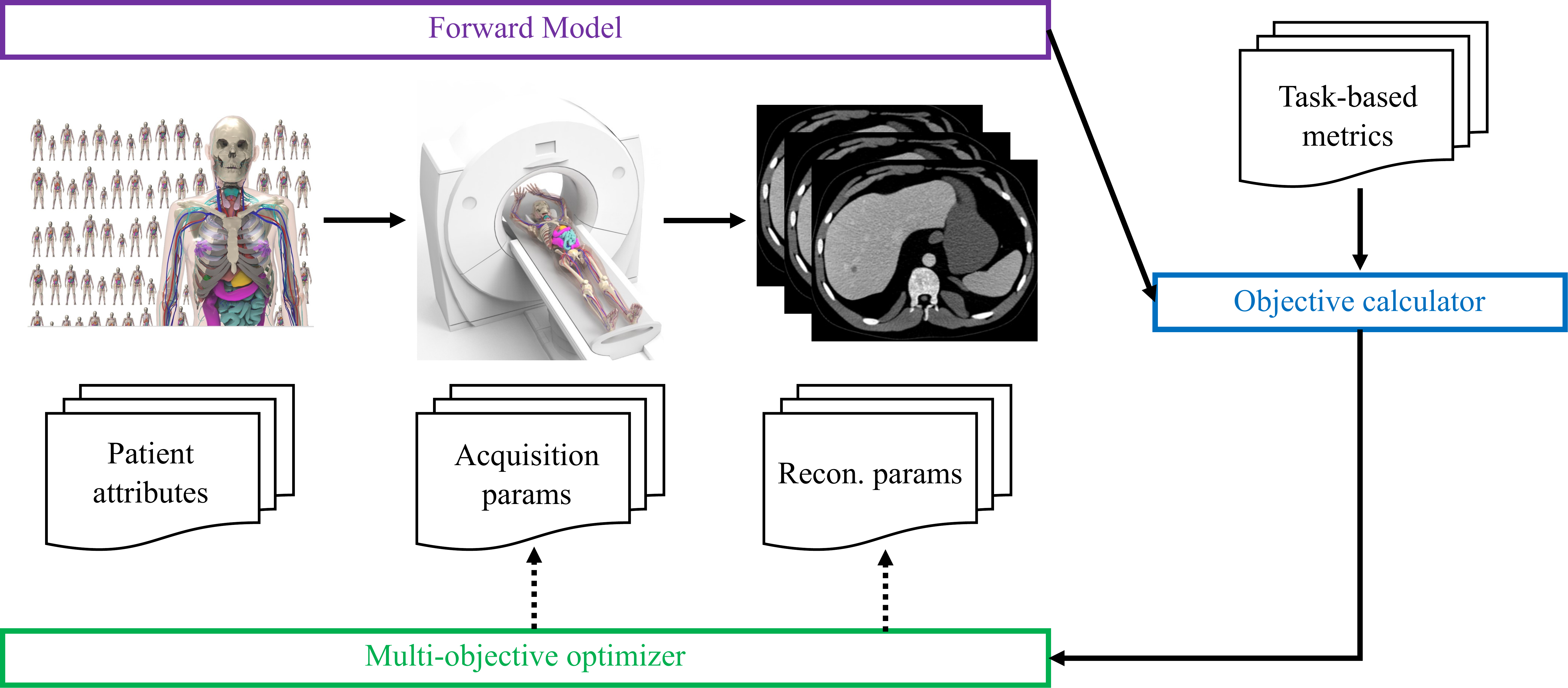}
    \caption{Outline of the optimization and Virtual Imaging Trial framework used to analyze the detectability index for liver lesions.}
    \label{fig:flowchart}
\end{figure*}
This study developed an optimization framework that integrates VITs with reinforcement learning to optimize CT imaging protocols. An overview of the framework is shown in Fig. \ref{fig:flowchart}. In summary, computational human models with liver lesions were imaged using a validated CT scanner model, and the resulting sinograms were reconstructed using an open-source reconstruction toolkit. The reconstructed images were analyzed to quantify image quality metrics, including noise, spatial resolution, and lesion contrast. These metrics were used to calculate the detectability index (d’) for the liver lesions. A reinforcement learning algorithm then served as the optimizer, using d’ as the objective function to identify parameter combinations that maximize detectability. Optimization performance was compared against an exhaustive search performed on a supercomputer.

\subsection{Computational Phantoms with Liver Lesions}
Fourteen (9 male, 5 female) human models (extended cardiac-torso, XCAT phantoms) were generated, each containing varying degrees of liver lesion severity \cite{RN15}.  The body habitus of these human models was based on real patient CT data, representing a cohort with a mean age of 51.3 +/- 13.7 years and a mean body mass index (BMI) of 26.5 +/- 5.7 kg/m\textsuperscript{2}. Each human model’s liver anatomy included detailed hepatic vasculature, generated using a physics-based vessel growth algorithm \cite{RN16}. One to six liver lesions of varying sizes (2.0 to 6.9 mm) were placed randomly within each liver. Further details on the development of these human models can be found in \cite{RN15}. 

\subsection{Image Acquisition}
The human models were imaged using a validated CT simulator (DukeSim) to generate projection images \cite{RN4,RN19,RN20,RN21}. DukeSim was set to model a vendor-generic scanner (Duke-1) with geometric and physical attributes that represent modern CT scanners \cite{RN15}. The acquisition parameter space included varied tube voltages (kV) and tube currents (mAs) that represent voltages and currents that are commonly used in clinical practice, summarized in Table \ref{tab:parameters}. The spectrum magnitude was adjusted per tube voltage to ensure similar flux for a given tube current. 

Due to the large number of simulations required for this study, the acquisitions were performed at the Oak Ridge National Laboratory using the Summit supercomputer which featured 4600 compute nodes. Each node contains 6 NVIDIA Tesla V100 GPUs and 2 IBM POWER9 CPUs per node plus 1TB of coherent memory \cite{RN27}.

\subsection{Image Reconstruction}
The acquired CT sinograms were reconstructed using the open-source Multi-Channel Reconstruction Toolkit (MCR Toolkit 2) \cite{RN5}. The reconstructions were performed using a filtered back projection technique where the filter shape is defined by the filter kernel and filter f50 which defines the equation and sharpness of the filter, respectively. The reconstruction parameters studied are summarized in Table \ref{tab:parameters}.

\begin{table}[ht]
\caption{Acquisition and reconstruction parameters}
\label{tab:parameters}
\centering

\renewcommand{\arraystretch}{1.3}
\setlength{\tabcolsep}{6pt}

\begin{tabular}{|>{\raggedright\arraybackslash}p{60pt}
                |>{\raggedright\arraybackslash}p{65pt}
                |>{\raggedright\arraybackslash}p{90pt}|}
\hline
\textbf{Parameter Type} & 
\textbf{Parameter Name} & 
\textbf{Parameter Space} \\\hline

\multirow[t]{2}{60pt}{Acquisition} & 
Tube Potential & 
100, 120, 140 kV \\\cline{2-3}
& Tube Current & 
25, 80, 150 mAs \\\hline

\multirow[t]{4}{60pt}{Reconstruction} & 
Filter Kernel & 
Ram-lak, Cosine, Smooth, Sharp, Enhancing \\\cline{2-3}
& Filter f50 & 
0.4, 0.6, 0.8 mm$^{-1}$ \\\cline{2-3}
& Slice Thickness & 
0.5, 1.0 mm \\\cline{2-3}
& Reconstructed Pixel Size & 
0.5, 1.0 mm \\\hline
\end{tabular}
\end{table}

\subsection{Image Analysis}
The image quality of the reconstructed images was evaluated using quantitative metrics that measure spatial resolution, noise, lesion contrast, and lesion detectability. 

Spatial resolution was quantified by calculating the modulation transfer function directly from the reconstructed images using an established methodology \cite{RN22}. Using this method, the skin-to-air interface was segmented and used to estimate the edge-spread function (ESF). The ESF was then differentiated to obtain the line-spread function (LSF) and then fit to a curve defined in \cite{RN23}. The Fourier transform of the LSF was calculated to obtain a CT resolution index (RI) which is an analog for the MTF. Finally, the frequency which corresponds to an intensity of 50\% of the maximum value was calculated and reported as MTF f\textsubscript{50}.

Image noise was calculated by selecting 100 random regions of interest (ROI) with CT numbers between -300 and 300 Hounsfield Units (HU). The standard deviation of the HU values within each ROI was measured, and the average value across all ROIs was computed as the global average noise index \cite{RN24}. Noise texture was calculated by measuring noise power spectrum (NPS) using the method defined in \cite{RN24}.

Lesion contrast was defined as the difference between the mean HU values of the surrounding liver parenchyma (\(\text{HU}_\text{liver}\)) and the lesion (\(\text{HU}_\text{lesion}\)). To measure it, lesions were segmented from the reconstructed images using the corresponding ground truth phantoms to guide the lesion locations. Each lesion mask was eroded with a 3x3 mm rectangle per slice to reduce uncertain measurements around the edges. \(\text{HU}_\text{lesion}\) was calculated as the average value within this eroded region. To compute \(\text{HU}_\text{liver}\), the lesion masks were dilated using a 25x25x6 mm cuboid, and the overlap with the liver parenchyma mask was used to extract the surrounding liver tissue voxels \cite{RN25}. The contrast-to-noise ratio (\(\text{CNR}\)) was also calculated for each lesion using:
\begin{align}
    \text{CNR} = \frac{\text{HU}_\text{liver} - \text{HU}_\text{lesion}}{\text{Noise}_\text{liver}}
\end{align}
where \(\text{Noise}_\text{liver}\) was the noise magnitude in the whole liver parenchyma and was obtained by segmenting the liver parenchyma, applying a 3x3x3 mm erosion to reduce the uncertain measurements around the edges, and computing the average standard deviation across all slices.

Lesion detectability was quantified using the detectability index (\(d'\)), a metric that reflects the likelihood of lesion detection by a human reader. For each lesion, \(d'\) was calculated in accordance with Smith \textit{et. al.,} \cite{RN26}:
\begin{align}
    d'^2=\frac{[\iint|W(u,v)|^2\cdot MTF^2(u,v)du\,dv]^2}{\iint|W(u,v)|^2\cdot MTF^2(u,v)\cdot NPS(u,v)du\,dv}
\end{align}
where \(W(u,v)\) is the Fourier transform of the task function defined by the lesion size and contrast, \(MTF(u,v)\) is the Modulation Transfer Function and \(NPS(u,v)\) is the Noise Power Spectrum. All components were defined in the spatial frequency domain, with \(u\) and \(v\) representing frequencies in the x and y directions, respectively. The task function assumed that the lesion is a circular disk and includes the diameter of the lesion and the contrast as measured earlier.

\subsection{Optimization Algorithm Development and Testing}
Several algorithm constraints were considered when selecting an appropriate optimization framework: (1) All actions (parameters) must be discrete, (2) the reward (detectability index) is positive and continuous, (3) the algorithm must be able to select combinations of multiple parameters as one action (multi-discrete), and (4) rewards are deterministic for a given action. Given these constraints, we selected a reinforcement learning approach because it replicates the traditional optimization approach using physical phantoms, and it uses a transparent and easily explainable decision-making process. For the reinforcement learning agent, a Proximal Policy Optimization (PPO) agent \cite{RN29} was selected since it meets the above constraints, is available in open-source packages \cite{RN30,RN31,RN32}, and is widely used in various applications, including alignment of large language models (LLMs) \cite{RN33,RN34}.

\begin{figure}[h]
    \centering
    \includegraphics[width=\linewidth]{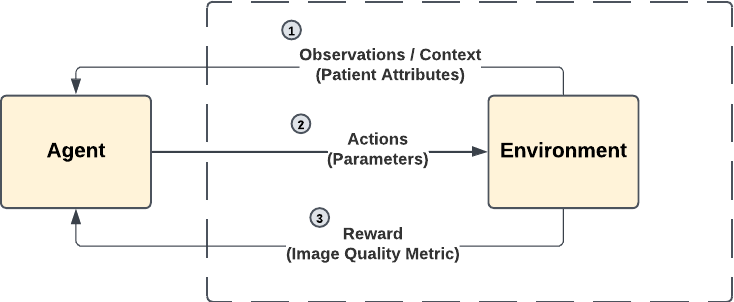}
    \caption{Block diagram of the reinforcement learning algorithm.}
    \label{fig:RL_block_diagram}
\end{figure}

The proposed reinforcement learning algorithm consisted of the PPO agent and an acquisition-reconstruction environment. The algorithm was used to determine the optimal parameters for each case. The input to the agent was a patient attribute vector, which included the patient’s BMI and sex, and a neural network architecture consisting of 2 layers with 64 units per layer. The network used hyperbolic tangent (Tanh) as the non-linearity function. The reward (to be maximized) was defined as the detectability index (\(d'\)) of the liver lesion located within the liver parenchyma of each phantom. A block diagram of the reinforcement learning algorithm is shown in Fig. \ref{fig:RL_block_diagram}.

The PPO agent was trained for 300 steps across 3 representative patients. For each step, the agent had the ability to acquire and reconstruct images for a single patient with any combination of parameters listed in Table \ref{tab:parameters}. The agent’s goal was to learn actions that maximize the reward, i.e., maximize \(d'\), for each episode. Although \(d'\) was used to define the reward in this study, it can be altered to other image quality metrics depending on the optimization task.

\subsection{Algorithm Analysis}
For each patient, an exhaustive search of all possible parameter combinations (as defined in Table \ref{tab:parameters}) was performed to establish a performance benchmark. The \(d'\) for each parameter combination was computed and used to analyze trends between the input parameters \(d'\). 

The accuracy of the developed optimizer was evaluated by comparing the \(d'\) obtained through the reinforcement learning approach to the maximum \(d'\) from the exhaustive search. The efficiency of the reinforcement learning model was assessed by comparing the number of steps required to achieve the maximum \(d'\) value using reinforcement learning with that of exhaustive search. 

\begin{figure*}[h]
    \centering
    \includegraphics[width=\linewidth]{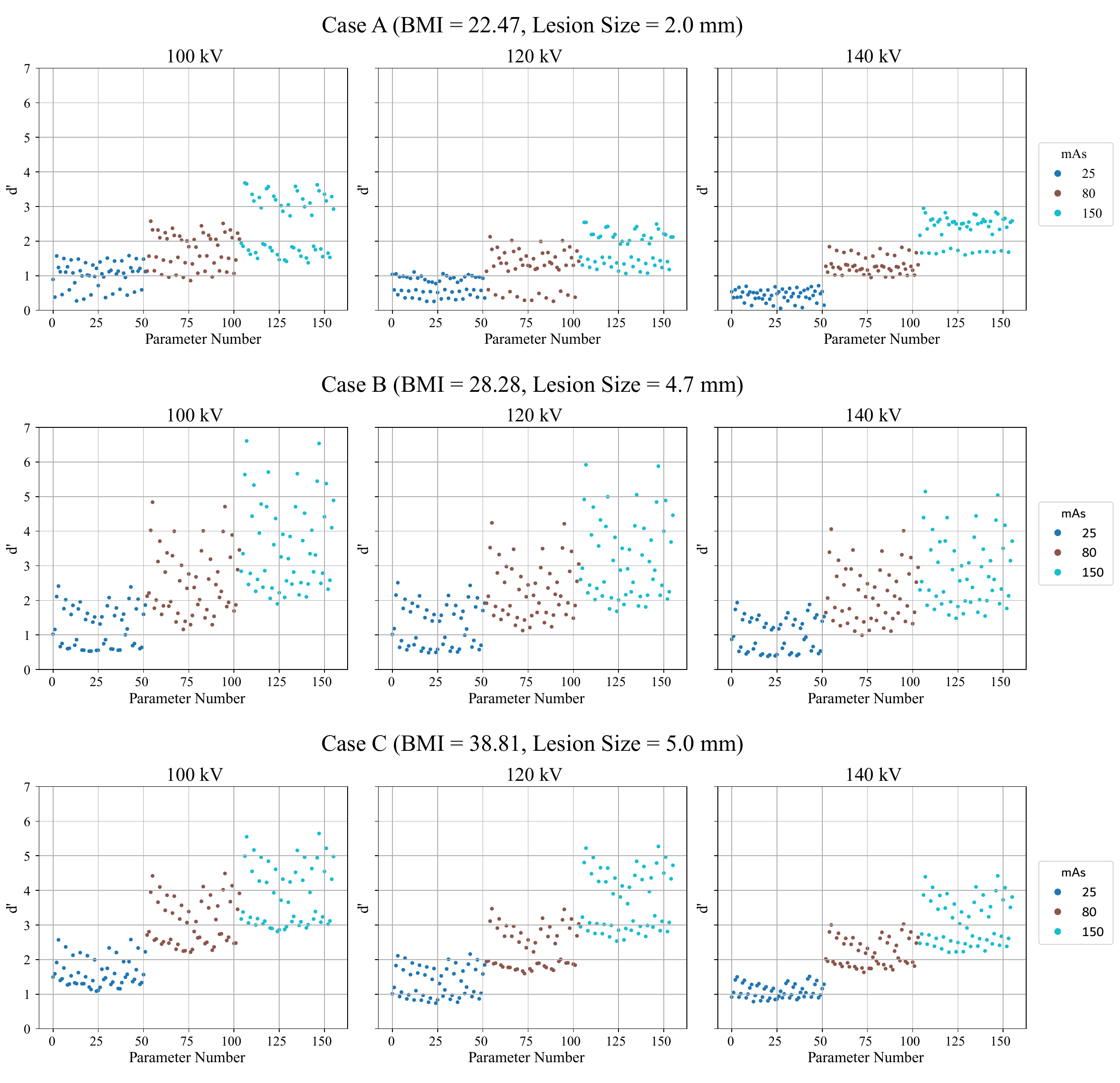}
    \caption{Exhaustive search results categorized by tube potential and tube current for three cases across all acquisition and reconstruction parameters defined in Table \ref{tab:parameters}.}
    \label{fig:exhaustive-search}
\end{figure*}

\section{Results}

\subsection{Exhaustive Search Results}

Fig. \ref{fig:exhaustive-search} shows the resulting \(d'\) value associated with the full parameter sweep across all combinations defined in Table \ref{tab:parameters} for three representative cases. These cases represent a range of BMI values (22.47, 28.28, and 38.81 kg/m\textsuperscript{2}) and lesion sizes (2.0, 4.7, and 5.0 mm). Each case involved 468 unique parameter combinations, accounting for variations in the acquisition and reconstruction parameters. The parameter combinations in each tube potential plot are ordered sequentially by tube current, filter kernel, filter f50, matrix size, and slice thickness such that each parameter combination number is unique and equal across all three cases. 

Across all cases, the exhaustive search results showed a consistent positive trend between tube current and \(d'\). These results also demonstrated considerable variability in \(d'\) values depending on the specific combination of acquisition and reconstruction parameters. The observed range of \(d'\) across the search space was 3.62, 6.23, and 4.90, for cases A, B, and C, respectively.

\begin{figure}[h]
    \centering
    \includegraphics[width=\linewidth]{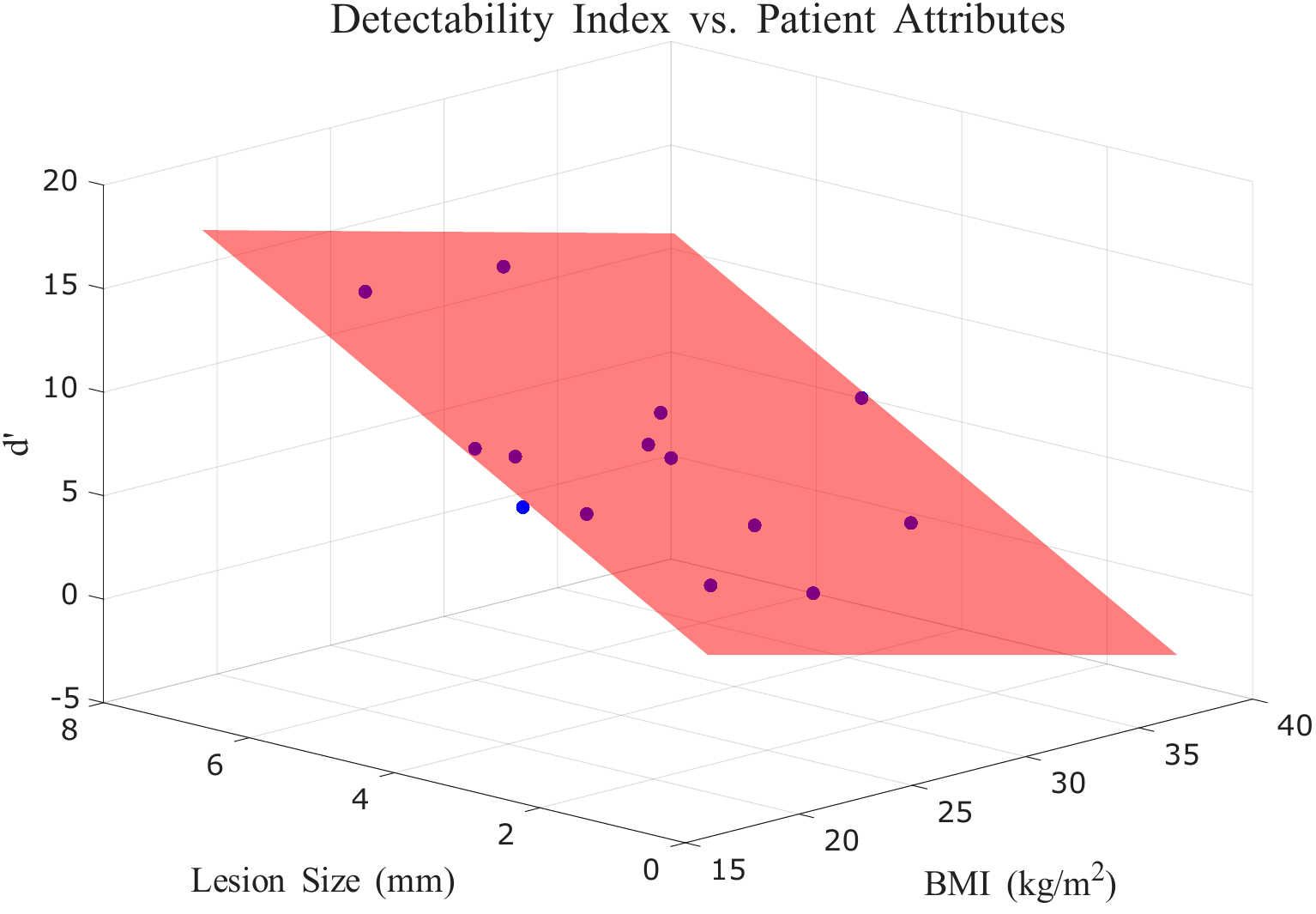}
    \caption{Maximum detectability index \(d'\) as a function of patient BMI and lesion size.}
    \label{fig:3d-plot}
\end{figure}

Fig. \ref{fig:3d-plot} shows the relationship between maximum \(d'\) and patient-specific attributes (BMI and lesion size) across all human models. A planar regression fit ($R^2=0.841$) was applied to model this relationship:
\begin{align}
   d'=-0.28\,\cdot\,\text{BMI}\,+\,2.10\,\cdot\,\text{Lesion\,Size}\,+5.51 
\end{align}
This result indicates that lesion detectability varies across patients as a function of both patient size and lesion size.

\begin{figure}[h!]
    \centering
    \includegraphics[width=\linewidth]{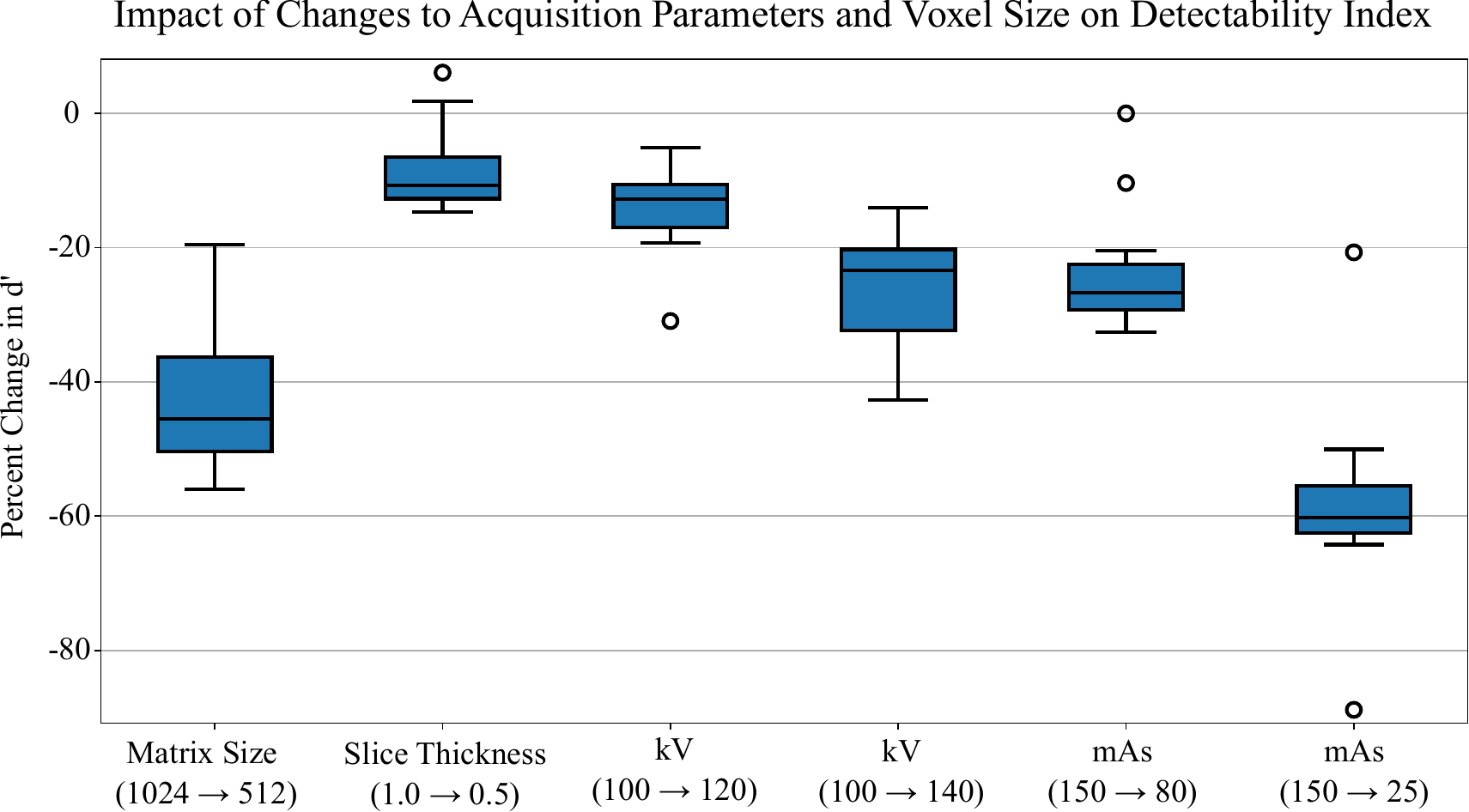}
    \caption{Impact of changes to acquisition parameters (tube potential and tube current) and voxel size (matrix size and slice thickness) on detectability index (\(d'\)).}
    \label{fig:acquisition-impact}
    \vspace{6pt}
    \includegraphics[width=\linewidth]{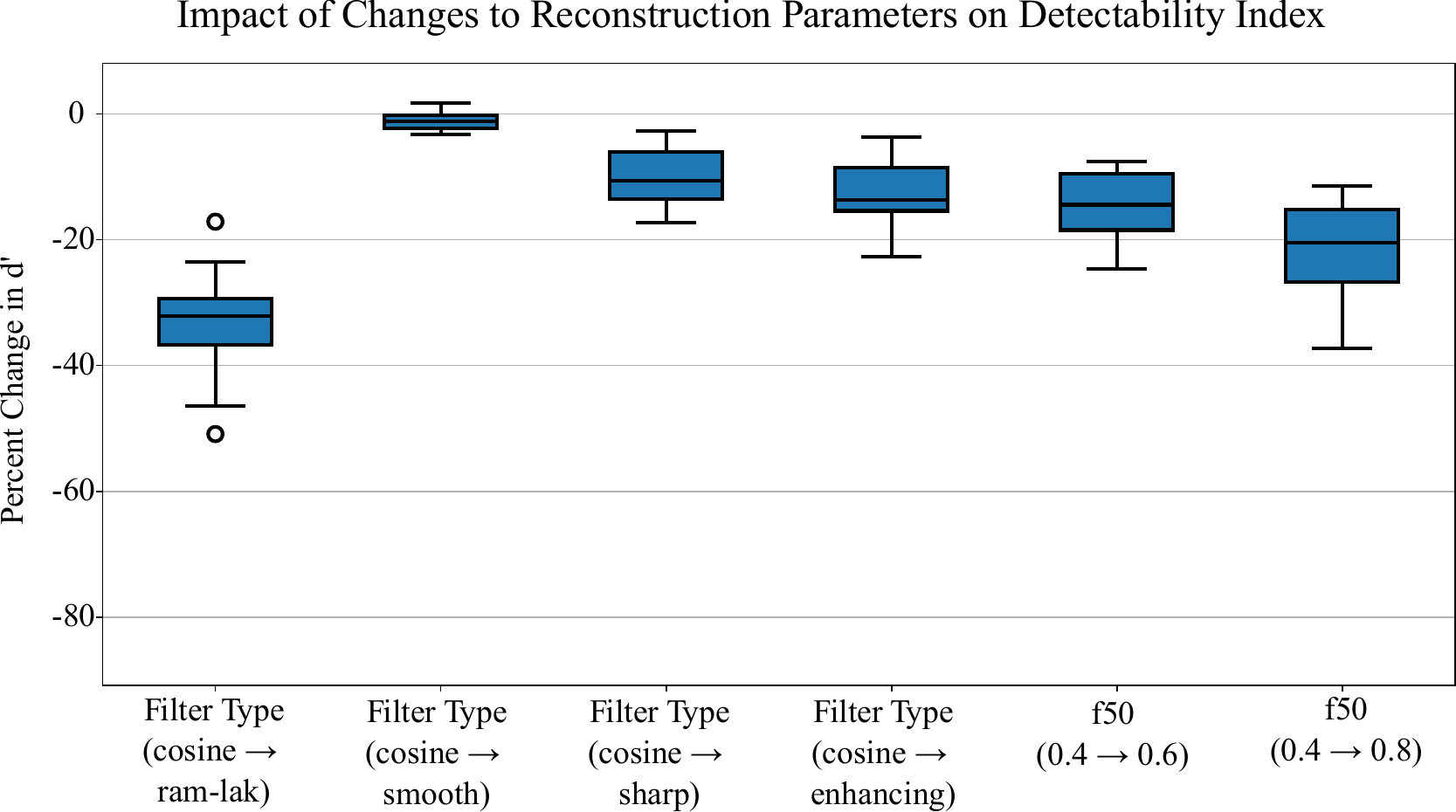}
    \caption{Impact of changes to reconstruction filter parameters (filter kernel and filter f50) to detectability index (\(d'\)).}
    \label{fig:reconstruction-impact}
\end{figure}

Further analysis showed that, for most cases, the optimal imaging conditions that maximized d’ corresponded to the maximum dose level (150 mAs), the minimum tube potential (100 kV), a cosine filter kernel, the minimum filter f50 (0.4), the maximum slice thickness (1.0 mm), and the minimum in-plane pixel size (1024). Using these default “optimal” conditions as a baseline, Fig. \ref{fig:acquisition-impact} and Fig. \ref{fig:reconstruction-impact} depict the percentage change in \(d'\) resulting from changing one parameter at a time. A negative change indicates that altering the parameter decreased the resulting \(d'\) whereas a positive change indicates changing the parameter improved \(d'\).

Overall, these results confirmed that matrix size and dose level had the largest impact on \(d'\). Filter kernel and slice thickness also influenced \(d'\), though with greater case-by-case variability.

\subsection{Qualitative Results}

\begin{figure}
    \centering
    \includegraphics[width=\linewidth]{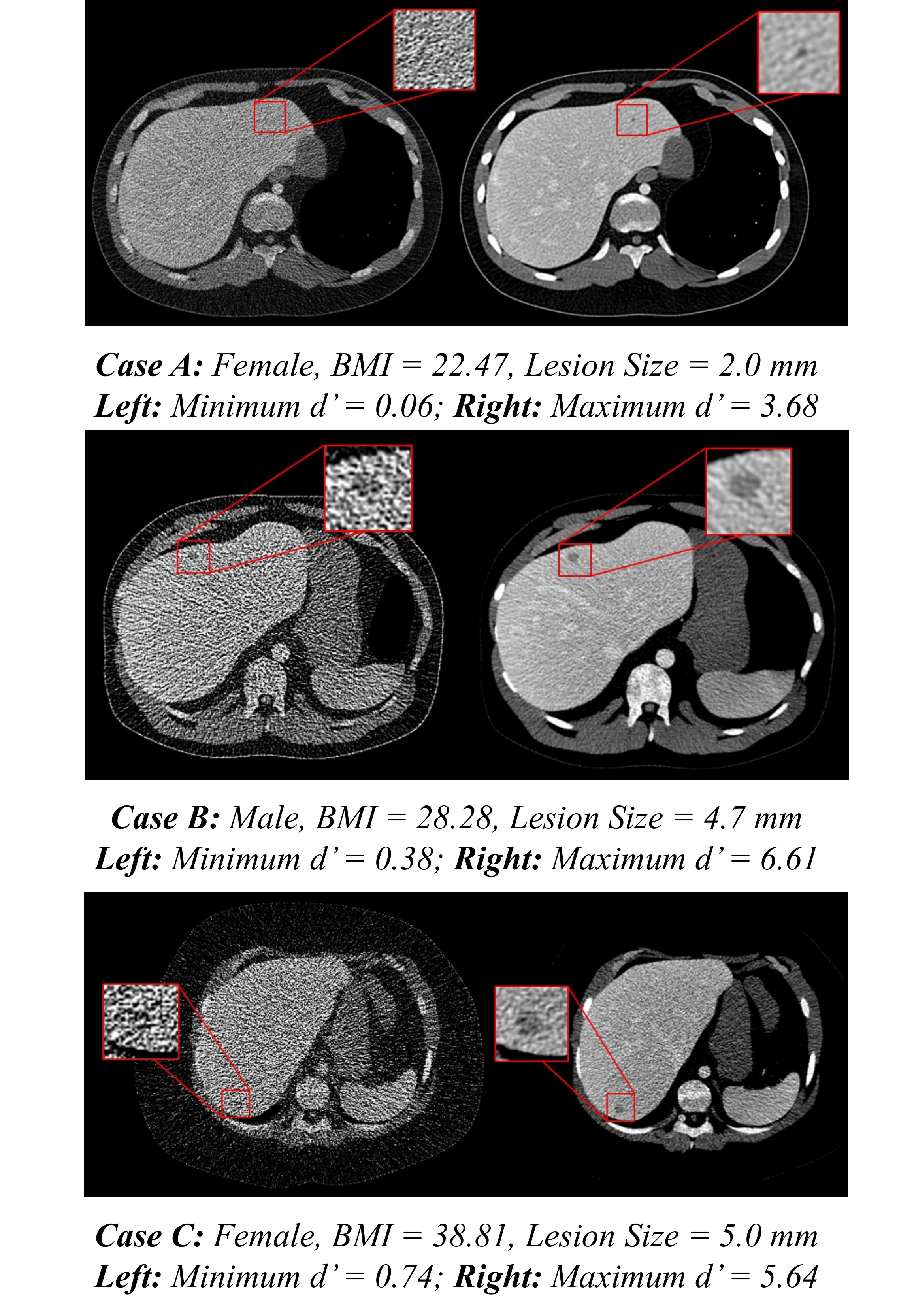}
    \caption{Qualitative comparison between \(d'\) values that correspond to the minimum and maximum values in the search space defined in Table \ref{tab:parameters}.}
    \label{fig:qualitative-images}
\end{figure}

Fig. \ref{fig:qualitative-images} provides a visual comparison of the reconstructed images that had the lowest and highest d’ values under the studied parameter space (Table \ref{tab:parameters}) for three representative liver lesion cases. Qualitatively, as the patient size increased from Cases A to C, indicated by patient BMI, noise magnitude also increased. Comparing Case A and B (similar BMIs, different lesion sizes), the larger lesion (4.7 mm vs. 2.0 mm) in Case B resulted in a higher detectability index (6.61 vs. 3.68), highlighting the influence of lesion size on detectability. In addition, Case A shows the importance of protocol optimization for a small lesion size. The unoptimized parameters on the left result in a \(d'\) close to 0 and a lesion that cannot be detected in the noisy image whereas the optimized parameters on the right result made the lesion appearance easier to detect with a \(d'\) of 3.68. Finally, Case C (BMI = 36.11) depicts the challenge of detecting lesions in larger patient.

\subsection{Reinforcement Learning Algorithm Performance}


\begin{figure}[t]
    \centering
    \includegraphics[trim=0in 0in 0in .5in, clip, width=\linewidth]{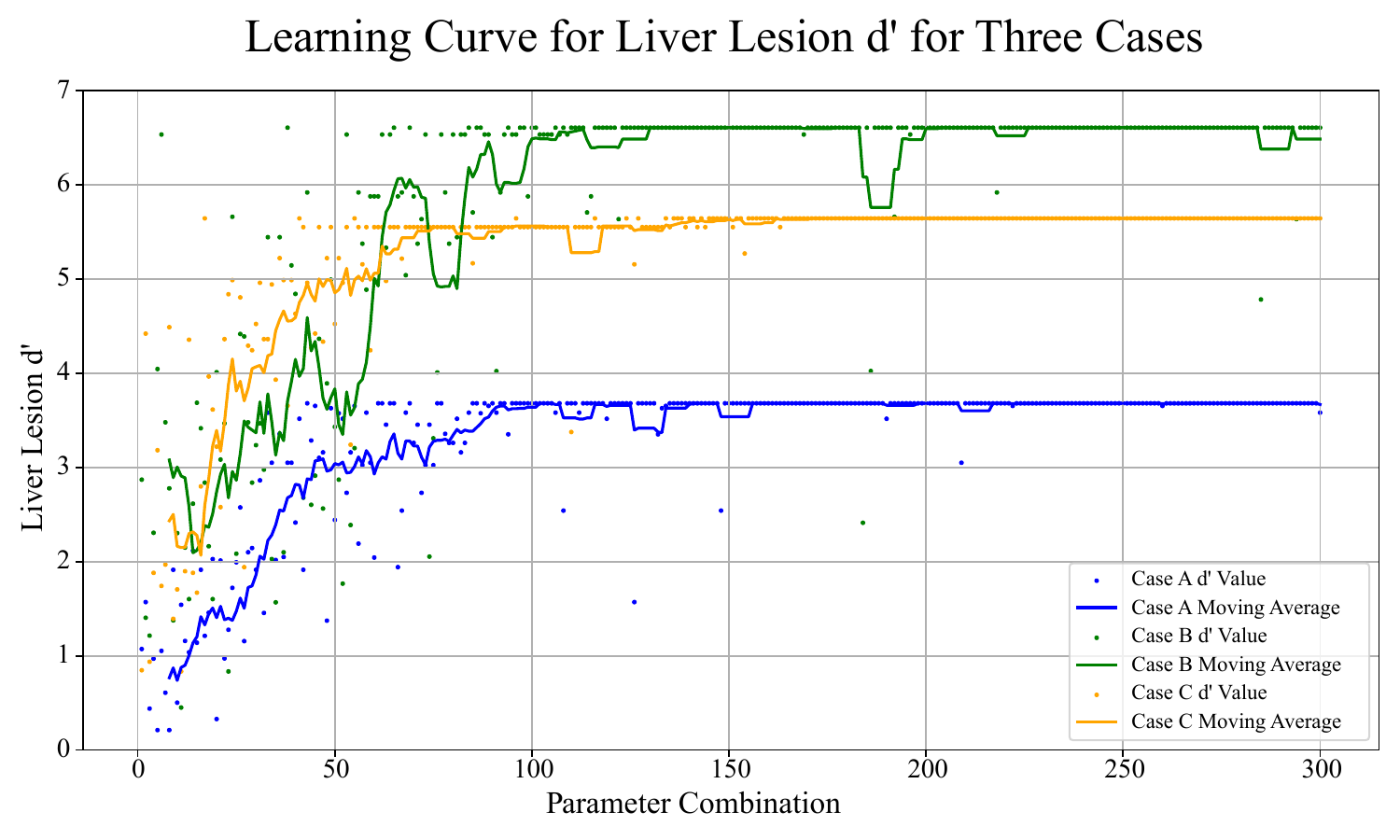}
    \caption{Learning curve for three liver lesion cases using the reinforcement learning algorithm. The algorithm interacts with the acquisition and reconstruction environment to determine the optimal parameters to achieve a maximum reward (\(d'\)).}
    \label{fig:learning-curve}
\end{figure}

Fig. \ref{fig:learning-curve} shows the learning curves for the reinforcement learning algorithm across three representative cases. In each case, the PPO agent interacted with the simulated acquisition-reconstruction environment to select parameter combinations that maximized the reward (liver lesion \(d'\)). 

For all cases, the reinforcement learning agent achieved the global maximum \(d'\) within approximately 100 parameter combinations, whereas the exhaustive search approach required 468 combinations. This demonstrates a 79.7\% reduction in steps needed to reach optimal setting. Additionally, because both the reinforcement learning and exhaustive search approaches identified the same maximum \(d'\) values, the reinforcement learning approach demonstrated zero error in optimization accuracy.

\section{Discussion}
This study leveraged a VIT framework and a reinforcement learning algorithm to optimize CT acquisition and reconstruction parameters for maximizing liver lesion detectability (\(d'\)) across a range of patient and lesion attributes. The exhaustive search indicated that \(d'\) was highest when using softer reconstruction kernels, lower tube potential, higher dose, and smaller in-plane pixel size.

In terms of acquisition parameters, increasing the tube current reduced image noise, thereby improving CNR and detectability index. Decreasing the tube potential (from 140 kV to 100 kV) generally increased the lesion contrast without increasing noise for a given tube current, leading to improved CNR and higher \(d'\).

With respect to reconstruction parameters, softer reconstruction kernels, which correspond to the smooth and cosine filter kernels and lower filter f50 values, consistently yielded the highest \(d'\). These kernels suppress image noise, thereby increasing CNR, albeit at the cost of spatial resolution. For low-contrast detection tasks, this tradeoff is beneficial, as reduced noise magnitude has a more significant positive impact on detectability than the modest loss in spatial resolution.

These findings are consistent with previous studies. Felice \textit{et. al.,} reported that softer reconstruction kernels provided the highest CNR and \(d'\) for energy integrating CT scans, and higher dose levels were also correlated with higher \(d'\) values \cite{RN25}. Similarly, Schindera \textit{et al.,} evaluated the effect of low tube voltages and high tube currents on CNR using a phantom methodology \cite{RN17}, showing that a decrease in tube voltage from 140 to 80 kV increased tumor lesion attenuation and the highest CNR was associated with the lowest tube voltage. Our results extend previous findings by evaluating such relationships within a high-dimensional parameter space using VITs.

Thanks to a superior balance between exploration and exploitation~\cite{del2024comparative}, the proposed reinforcement learning optimizer proved highly effective in navigating the complex parameter space in this study and finding the maximum \(d'\) value in fewer steps than an exhaustive search approach. 

An important future application of this optimizer is to determine which dose level is required to achieve a minimum acceptable \(d'\) threshold. This threshold can be set by converting \(d'\) to detection accuracy, \(A_d\), using the equation and methodology presented in Solomon et. al.  \cite{RN18}:
\begin{align}
    A_x=\Phi\left(\frac{x}{\sqrt{2}}\right)
\end{align}
where \(x\) denotes the image quality metric of interest, \(d'\), and \(\Phi(\cdot)\) represents the normal cumulative distribution function. By setting \(A_x\) to a high detection accuracy threshold (e.g., 99\%) and using a calibration factor acquired from a reader study, one could calculate a corresponding \(d'\) threshold that meets this detection accuracy threshold. For example, in Case B (Fig. \ref{fig:qualitative-images}), the maximum \(d'\) (6.61) corresponded with the highest dose level of 150 mAs. However, if the acceptable \(d'\) threshold was less than 6.1, then the algorithm would determine that a lower dose level is the optimal setting. This approach has potential clinical benefits by ensuring that the radiation dose received by patients is minimized while meeting adequate image quality requirements for the clinical task. 

While the proposed optimization framework demonstrated promising performance and several future applications, this study had some limitations. First, the optimizer was targeted to maximize a single objective and clinical task, \(d'\), and did not include other quality metrics. Given the modularity of the agent/environment, the same proposed framework could be extended to include multi-objective optimization addressing a clinical task at hand. Additionally, the algorithm was tested using a limited number of human models with liver lesions. Expanding the patient cohort may enable better characterization of inter-patient variability and more robust conclusions regarding optimal protocol settings across the target population by enhancing the generalization capability of the reinforcement learning agent to patients unseen during the training process.

Lastly, the current reinforcement learning reward function was solely based on maximizing image quality (\(d'\)), without penalizing radiation dose. Future work may incorporate dose into the reward function (e.g., by introducing a dose penalty or weighted trade-off) to better reflect the clinical objective of balancing between image quality and radiation dose.

\section{Conclusion}

This study demonstrated that integrating virtual imaging tools and a reinforcement learning-based optimizer can efficiently and accurately identify optimal CT acquisition and reconstruction parameters for maximizing lesion detectability, outperforming traditional exhaustive search methods in terms of computational efficiency. The proposed framework is adaptable and extensible, enabling optimization across diverse imaging parameter spaces, image quality metrics, and clinical tasks. As such, it holds great potential for advancing personalized, task-specific imaging protocol design in clinical practice. 

\begin{credits}
\bigskip
\textbf{\ackname} This work was supported in part by the National Institutes of Health (R01HL155293 and P41EB028744). This manuscript has been authored in part by UT-Battelle, LLC, under contract DE-AC05-00OR22725 with the US Department of Energy (DOE). The US government retains and the publisher, by accepting the article for publication, acknowledges that the US government retains a nonexclusive, paid-up, irrevocable, worldwide license to publish or reproduce the published form of this manuscript, or allow others to do so, for US government purposes. DOE will provide public access to these results of federally sponsored research in accordance with the DOE Public Access Plan (\href{http://energy.gov/downloads/doe-public-access-plan}{http://energy.gov/downloads/doe-public-access-plan}).
\end{credits}

\printbibliography

@article{RN1,
    key = {AAPM Protocol Management},
   title = {AAPM Medical Physics Practice Guideline 1.a: CT Protocol Management and Review Practice Guideline},
   journal = {Journal of Applied Clinical Medical Physics},
   volume = {14},
   number = {5},
   pages = {3–12},
   ISSN = {1526-9914},
   DOI = {https://doi.org/10.1120/jacmp.v14i5.4462},
   url = {https://aapm.onlinelibrary.wiley.com/doi/abs/10.1120/jacmp.v14i5.4462},
   year = {2013},
   type = {Journal Article}
}

@article{RN20,
   author = {Abadi, E. and Harrawood, B. and Rajagopal, J. R. and Sharma, S. and Kapadia, A. and Segars, W. P. and Stierstorfer, K. and Sedlmair, M. and Jones, E. and Samei, E.},
   title = {Development of a scanner-specific simulation framework for photon-counting computed tomography},
   journal = {Biomed Phys Eng Express},
   volume = {5},
   number = {5},
   ISSN = {2057-1976},
   DOI = {10.1088/2057-1976/ab37e9},
   year = {2019},
   type = {Journal Article}
}

@article{RN4,
   author = {Abadi, E. and Harrawood, B. and Sharma, S. and Kapadia, A. and Segars, W. P. and Samei, E.},
   title = {DukeSim: A Realistic, Rapid, and Scanner-Specific Simulation Framework in Computed Tomography},
   journal = {IEEE Trans Med Imaging},
   volume = {38},
   number = {6},
   pages = {1457–1465},
   ISSN = {1558-254X (Electronic)
0278-0062 (Print)
0278-0062 (Linking)},
   DOI = {10.1109/TMI.2018.2886530},
   url = {https://www.ncbi.nlm.nih.gov/pubmed/30561344},
   year = {2019},
   type = {Journal Article}
}

@article{RN15,
   author = {Abadi, E. and Segars, W. P. and Felice, N. and Sotoudeh-Paima, S. and Hoffman, E. A. and Wang, X. and Wang, W. and Clark, D. and Ye, S. and Jadick, G. and Fryling, M. and Frush, D. P. and Samei, E.},
   title = {AAPM Truth-based CT (TrueCT) reconstruction grand challenge},
   journal = {Med Phys},
   ISSN = {0094-2405},
   DOI = {10.1002/mp.17619},
   year = {2025},
   type = {Journal Article}
}

@article{RN13,
   author = {Abadi, E. and Segars, W. P. and Harrawood, B. and Sharma, S. and Kapadia, A. and Samei, E.},
   title = {Virtual clinical trial for quantifying the effects of beam collimation and pitch on image quality in computed tomography},
   journal = {J Med Imaging (Bellingham)},
   volume = {7},
   number = {4},
   pages = {042806},
   ISSN = {2329-4302 (Print)
2329-4310 (Electronic)
2329-4302 (Linking)},
   DOI = {10.1117/1.JMI.7.4.042806},
   url = {https://www.ncbi.nlm.nih.gov/pubmed/32509918},
   year = {2020},
   type = {Journal Article}
}

@article{RN14,
   author = {Abadi, E. and Segars, W. P. and Tsui, B. M. W. and Kinahan, P. E. and Bottenus, N. and Frangi, A. F. and Maidment, A. and Lo, J. and Samei, E.},
   title = {Virtual clinical trials in medical imaging: a review},
   journal = {J Med Imaging (Bellingham)},
   volume = {7},
   number = {4},
   pages = {042805},
   ISSN = {2329-4302 (Print)
2329-4302},
   DOI = {10.1117/1.Jmi.7.4.042805},
   year = {2020},
   type = {Journal Article}
}

@article{RN11,
   author = {Barufaldi, B. and Maidment, A. D. A. and Dustler, M. and Axelsson, R. and Tomic, H. and Zackrisson, S. and Tingberg, A. and Bakic, P. R.},
   title = {Virtual Clinical Trials in Medical Imaging System Evaluation and Optimisation},
   journal = {Radiat Prot Dosimetry},
   volume = {195},
   number = {3-4},
   pages = {363–371},
   ISSN = {1742-3406 (Electronic)
0144-8420 (Print)
0144-8420 (Linking)},
   DOI = {10.1093/rpd/ncab080},
   url = {https://www.ncbi.nlm.nih.gov/pubmed/34144597},
   year = {2021},
   type = {Journal Article}
}

@article{RN31,
   author = {Bou, Albert and Bettini, Matteo and Dittert, Sebastian and Kumar, Vikash and Sodhani, Shagun and Yang, Xiaomeng and De Fabritiis, Gianni and Moens, Vincent},
   title = {Torchrl: A data-driven decision-making library for pytorch},
   journal = {arXiv preprint arXiv:2306.00577},
   year = {2023},
   type = {Journal Article}
}

@article{RN24,
   author = {Christianson, Olav and Winslow, James and Frush, Donald P. and Samei, Ehsan},
   title = {Automated Technique to Measure Noise in Clinical CT Examinations},
   journal = {American Journal of Roentgenology},
   volume = {205},
   number = {1},
   pages = {W93–W99},
   DOI = {10.2214/ajr.14.13613},
   url = {https://ajronline.org/doi/abs/10.2214/AJR.14.13613},
   year = {2015},
   type = {Journal Article}
}

@article{RN5,
   author = {Clark, D. P. and Badea, C. T.},
   title = {MCR toolkit: A GPU-based toolkit for multi-channel reconstruction of preclinical and clinical x-ray CT data},
   journal = {Med Phys},
   volume = {50},
   number = {8},
   pages = {4775–4796},
   ISSN = {2473-4209 (Electronic)
0094-2405 (Print)
0094-2405 (Linking)},
   DOI = {10.1002/mp.16532},
   url = {https://www.ncbi.nlm.nih.gov/pubmed/37285215},
   year = {2023},
   type = {Journal Article}
}

@article{RN25,
   author = {Felice, Nicholas and Wildman-Tobriner, Benjamin and Segars, William and Bashir, Mustafa and Marin, Daniele and Samei, Ehsan and Abadi, Ehsan},
   title = {Photon-counting computed tomography versus energy-integrating computed tomography for detection of small liver lesions: comparison using a virtual framework imaging},
   journal = {Journal of Medical Imaging},
   volume = {11},
   number = {5},
   pages = {053502},
   url = {https://doi.org/10.1117/1.JMI.11.5.053502},
   year = {2024},
   type = {Journal Article}
}

@article{RN33,
   author = {Havrilla, Alex and Du, Yuqing and Raparthy, Sharath Chandra and Nalmpantis, Christoforos and Dwivedi-Yu, Jane and Zhuravinskyi, Maksym and Hambro, Eric and Sukhbaatar, Sainbayar and Raileanu, Roberta},
   title = {Teaching large language models to reason with reinforcement learning},
   journal = {arXiv preprint arXiv:2403.04642},
   year = {2024},
   type = {Journal Article}
}

@article{RN34,
   author = {Hu, Bokai and Somayajula, Sai Ashish and Pan, Xin and Xie, Pengtao},
   title = {Improving the Language Understanding Capabilities of Large Language Models Using Reinforcement Learning},
   journal = {arXiv preprint arXiv:2410.11020},
   year = {2024},
   type = {Journal Article}
}

@article{RN35,
   author = {Kaelbling, Leslie Pack and Littman, Michael L and Moore, Andrew W},
   title = {Reinforcement learning: A survey},
   journal = {Journal of artificial intelligence research},
   volume = {4},
   pages = {237–285},
   ISSN = {1076-9757},
   year = {1996},
   type = {Journal Article}
}

@article{RN2,
   author = {Martini, K. and Moon, J. W. and Revel, M. P. and Dangeard, S. and Ruan, C. and Chassagnon, G.},
   title = {Optimization of acquisition parameters for reduced-dose thoracic CT: A phantom study},
   journal = {Diagn Interv Imaging},
   volume = {101},
   number = {5},
   pages = {269–279},
   ISSN = {2211-5684 (Electronic)
2211-5684 (Linking)},
   DOI = {10.1016/j.diii.2020.01.012},
   url = {https://www.ncbi.nlm.nih.gov/pubmed/32107196},
   year = {2020},
   type = {Journal Article}
}

@article{RN19,
   author = {McCabe, Cindy and Harrawood, Brian and Samei, Ehsan and Abadi, Ehsan},
   title = {In silico modeling of a clinical photon-counting CT system: Verification and validation},
   journal = {Medical Physics},
   volume = {52},
   number = {6},
   pages = {3840–3853},
   ISSN = {0094-2405},
   DOI = {https://doi.org/10.1002/mp.17886},
   url = {https://aapm.onlinelibrary.wiley.com/doi/abs/10.1002/mp.17886},
   year = {2025},
   type = {Journal Article}
}

@article{RN23,
   author = {Ott, Julien G. and Becce, Fabio and Monnin, Pascal and Schmidt, Sabine and Bochud, François O. and Verdun, Francis R.},
   title = {Update on the non-prewhitening model observer in computed tomography for the assessment of the adaptive statistical and model-based iterative reconstruction algorithms},
   journal = {Physics in Medicine \& Biology},
   volume = {59},
   number = {15},
   pages = {4047},
   ISSN = {0031-9155},
   DOI = {10.1088/0031-9155/59/4/4047},
   url = {https://dx.doi.org/10.1088/0031-9155/59/4/4047},
   year = {2014},
   type = {Journal Article}
}

@article{RN30,
   author = {Raffin, Antonin and Hill, Ashley and Gleave, Adam and Kanervisto, Anssi and Ernestus, Maximilian and Dormann, Noah},
   title = {Stable-baselines3: Reliable reinforcement learning implementations},
   journal = {Journal of machine learning research},
   volume = {22},
   number = {268},
   pages = {1–8},
   ISSN = {1533-7928},
   year = {2021},
   type = {Journal Article}
}

@article{RN22,
   author = {Sanders, Jeremiah and Hurwitz, Lynne and Samei, Ehsan},
   title = {Patient-specific quantification of image quality: An automated method for measuring spatial resolution in clinical CT images},
   journal = {Medical Physics},
   volume = {43},
   number = {10},
   pages = {5330–5338},
   ISSN = {0094-2405},
   DOI = {https://doi.org/10.1118/1.4961984},
   url = {https://aapm.onlinelibrary.wiley.com/doi/abs/10.1118/1.4961984},
   year = {2016},
   type = {Journal Article}
}

@article{RN16,
   author = {Sauer, T. J. and Bejan, A. and Segars, P. and Samei, E.},
   title = {Development and CT image-domain validation of a computational lung lesion model for use in virtual imaging trials},
   journal = {Med Phys},
   volume = {50},
   number = {7},
   pages = {4366–4378},
   ISSN = {0094-2405 (Print)
0094-2405},
   DOI = {10.1002/mp.16222},
   year = {2023},
   type = {Journal Article}
}

@article{RN17,
   author = {Schindera, Sebastian T. and Nelson, Rendon C. and Mukundan, Srinivasan and Paulson, Erik K. and Jaffe, Tracy A. and Miller, Chad M. and DeLong, David M. and Kawaji, Keigo and Yoshizumi, Terry T. and Samei, Ehsan},
   title = {Hypervascular Liver Tumors: Low Tube Voltage, High Tube Current Multi–Detector Row CT for Enhanced Detection—Phantom Study},
   journal = {Radiology},
   volume = {246},
   number = {1},
   pages = {125–132},
   DOI = {10.1148/radiol.2461070307},
   url = {https://pubs.rsna.org/doi/abs/10.1148/radiol.2461070307},
   year = {2008},
   type = {Journal Article}
}

@article{RN29,
   author = {Schulman, John and Wolski, Filip and Dhariwal, Prafulla and Radford, Alec and Klimov, Oleg},
   title = {Proximal policy optimization algorithms},
   journal = {arXiv preprint arXiv:1707.06347},
   year = {2017},
   type = {Journal Article}
}

@article{RN32,
   author = {Serrano-Munoz, Antonio and Chrysostomou, Dimitrios and Bøgh, Simon and Arana-Arexolaleiba, Nestor},
   title = {skrl: Modular and flexible library for reinforcement learning},
   journal = {Journal of Machine Learning Research},
   volume = {24},
   number = {254},
   pages = {1–9},
   ISSN = {1533-7928},
   year = {2023},
   type = {Journal Article}
}

@article{RN21,
   author = {Shankar, Sachin S. and Felice, Nicholas and Hoffman, Eric A. and Atha, Jarron and Sieren, Jessica C. and Samei, Ehsan and Abadi, Ehsan},
   title = {Task-based validation and application of a scanner-specific CT simulator using an anthropomorphic phantom},
   journal = {Medical Physics},
   volume = {49},
   number = {12},
   pages = {7447–7457},
   ISSN = {0094-2405},
   DOI = {https://doi.org/10.1002/mp.15967},
   url = {https://aapm.onlinelibrary.wiley.com/doi/abs/10.1002/mp.15967},
   year = {2022},
   type = {Journal Article}
}

@article{RN26,
   author = {Smith, Taylor Brunton and Solomon, Justin and Samei, Ehsan},
   title = {Estimating detectability index in vivo: development and validation of an automated methodology},
   journal = {Journal of Medical Imaging},
   volume = {5},
   number = {3},
   pages = {031403},
   url = {https://doi.org/10.1117/1.JMI.5.3.031403},
   year = {2017},
   type = {Journal Article}
}

@article{RN18,
   author = {Solomon, Justin and Samei, Ehsan},
   title = {Correlation between human detection accuracy and observer model-based image quality metrics in computed tomography},
   journal = {Journal of Medical Imaging},
   volume = {3},
   number = {3},
   pages = {035506},
   url = {https://doi.org/10.1117/1.JMI.3.3.035506},
   year = {2016},
   type = {Journal Article}
}

@article{RN12,
   author = {Vancoillie, L. and Cockmartin, L. and Lueck, F. and Marshall, N. and Keupers, M. and Nanke, R. and Kappler, S. and Van Ongeval, C. and Bosmans, H.},
   title = {Optimized signal of calcifications in wide-angle digital breast tomosynthesis: a virtual imaging trial},
   journal = {Eur Radiol},
   ISSN = {1432-1084 (Electronic)
0938-7994 (Linking)},
   DOI = {10.1007/s00330-024-10712-9},
   url = {https://www.ncbi.nlm.nih.gov/pubmed/38546790},
   year = {2024},
   type = {Journal Article}
}

@misc{RN27,
   author = {Wells, Jack and Bland, Buddy and Nichols, Jeff and Hack, Jim and Foertter, Fernanda and Hagen, Gaute and Maier, Thomas and Ashfaq, Moetasim and Messer, Bronson and Parete-Koon, Suzanne},
   title = {Announcing Supercomputer Summit},
   publisher = {Oak Ridge National Laboratory (ORNL), Oak Ridge, TN (United States)},
   number = {None},
   url = {https://www.osti.gov/biblio/1259664
https://www.osti.gov/servlets/purl/1259664},
   year = {2016},
   type = {Audiovisual Material}
}

@article{RN36,
   author = {Wiering, Marco A and Van Otterlo, Martijn},
   title = {Reinforcement learning},
   journal = {Adaptation, learning, and optimization},
   volume = {12},
   number = {3},
   pages = {729},
   year = {2012},
   type = {Journal Article}
}

@article{RN3,
   author = {Zarb, Francis and Rainford, Louise and McEntee, Mark F.},
   title = {Developing optimized CT scan protocols: Phantom measurements of image quality},
   journal = {Radiography},
   volume = {17},
   number = {2},
   pages = {109–114},
   year = {2011},
   type = {Journal Article}
}

@ARTICLE{del2024comparative,
  author={Rio, Alberto del and Jimenez, David and Serrano, Javier},
  journal={IEEE Access}, 
  title={Comparative Analysis of A3C and PPO Algorithms in Reinforcement Learning: A Survey on General Environments}, 
  year={2024},
  volume={12},
  number={},
  pages={146795-146806},
  doi={10.1109/ACCESS.2024.3472473}}

@inproceedings{fenwick2025,
  title={Black-box optimization of CT acquisition and reconstruction parameters: a reinforcement learning approach},
  author={Fenwick, David and NaderiAlizadeh, Navid and Tarokh, Vahid and Clark, Darin and Rajagopal, Jayasai and Kapadia, Anuj and Felice, Nicholas and Samei, Ehsan and Abadi, Ehsan},
  booktitle={Medical Imaging 2025: Physics of Medical Imaging},
  volume={13405},
  pages={463--467},
  year={2025},
  organization={SPIE}
}

\end{document}